\documentclass{article}

\usepackage{arxiv}

\usepackage[utf8]{inputenc} 
\usepackage[T1]{fontenc}    
\usepackage[colorlinks=true,linkcolor=blue,anchorcolor=black,citecolor=black,filecolor=black,menucolor=black,runcolor=black,urlcolor=black]{hyperref}       
\usepackage{url}            
\usepackage{booktabs}       
\usepackage{amsfonts}       
\usepackage{nicefrac}       
\usepackage{microtype}      
\usepackage{lipsum}		
\usepackage{graphicx}
\usepackage{natbib}
\usepackage{doi}
\usepackage{amsmath}

\title{PathFinder: Joint Decompositions of Linked Multimodal Datasets}

\author{ 
    \href{https://orcid.org/0000-0000-0000-0000}{\includegraphics[scale=0.06]{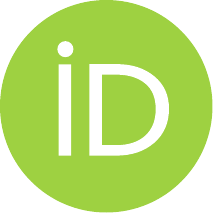}\hspace{1mm}Ying-Qiu Zheng}\thanks{Corresponding Author} \\
	Oxford Centre for Integrative Neuroimaging\\
	University of Oxford\\
	Oxford, UK \\
	\texttt{ying-qiu.zheng@ndcn.ox.ac.uk} \\
	\And
	\href{https://orcid.org/0000-0000-0000-0000}{\includegraphics[scale=0.06]{orcid.pdf}\hspace{1mm}Alex Fung} \\
	Oxford Centre for Integrative Neuroimaging\\
	University of Oxford\\
	Oxford, UK \\
	\texttt{alex.fung@ndcn.ox.ac.uk} \\
	\And
	\href{https://orcid.org/0000-0000-0000-0000}{\includegraphics[scale=0.06]{orcid.pdf}\hspace{1mm}Stephen M Smith} \\
	Oxford Centre for Integrative Neuroimaging\\
	University of Oxford\\
	Oxford, UK \\
	\texttt{stephen.smith@ndcn.ox.ac.uk} \\
    \And
	\href{https://orcid.org/0000-0000-0000-0000}{\includegraphics[scale=0.06]{orcid.pdf}\hspace{1mm}Rogier B Mars} \\
	Oxford Centre for Integrative Neuroimaging\\
	University of Oxford\\
	Oxford, UK \\
	\texttt{rogier.mars@ndcn.ox.ac.uk} \\
    \And
    \href{https://orcid.org/0000-0003-3234-5639}{\includegraphics[scale=0.06]{orcid.pdf}\hspace{1mm}Saad Jbabdi} \\
	Oxford Centre for Integrative Neuroimaging\\
	University of Oxford\\
	Oxford, UK \\
	\texttt{saad.jbabdi@ndcn.ox.ac.uk} \\
}

\renewcommand{\shorttitle}{\textit{PathFinder}: Joint Decompositions of Linked Multimodal Datasets}

\hypersetup{
pdftitle={A template for the arxiv style},
pdfsubject={q-bio.NC, q-bio.QM},
pdfauthor={David S.~Hippocampus, Elias D.~Striatum},
pdfkeywords={First keyword, Second keyword, More},
}

\begin{document}
\maketitle

\begin{abstract}
	Low-rank matrix decompositions can uncover patterns and structure in data and have a number of different applications across many disciplines. Extensions to "joint" low-rank decompositions have been proposed to link datasets from different modalities. While these methods enable the discovery of common patterns across modalities, they require that all the multimodal data share one or more dimensions. We propose a new analysis method, PathFinder, that enables co-analysis of datasets that do not necessarily all share a dimension. The key insight is that as long as pairs or subgroups of matrices do share some dimension, and that there are one or more paths that link across the data matrices, a global joint decomposition can be sought out. This enables the joint estimation of common patterns across different modalities, species, or scales, where a one-to-one mapping across all data along some dimension is not necessarily available. We show that PathFinder is a general umbrella under which many matrix decomposition methods fall as special cases. It can be used to discover common patterns across disparate datasets and to make predictions for missing data or modalities.
\end{abstract}


\section{Introduction}
Low-rank matrix decomposition methods have had a significant impact in neuroimaging. Being purely data-driven, they can reveal hidden structure in data where we lack a forward model, such as in time series linear models in task FMRI. By far the most popular matrix decomposition method is independent component analysis (ICA), which is widely used to map resting state brain networks at both single-subject and group levels \citep{Beckmann2004,Beckmann2005,Smith2009} , or to isolate and remove structured MRI \citep{Pruim2015,SalimiKhorshidi2014,Griffanti2017} or EEG artefacts \citep{Jung2000,Delorme2007}. Several extensions to ICA have been proposed to analyse multiple subjects or modalities simultaneously. Examples include concatenation \citep{Calhoun2001}, tensor-ICA \citep{Beckmann2005} and linked-ICA \citep{Groves2011}. Many other methods for joint analyses of datasets also exist \citep{Singh2008,Zitnik2015} , see \citep{Sui2012} for a survey. What these joint decomposition methods have in common is that they enable the discovery of patterns that are shared across datasets. In all these methods, the input data matrices must share at least one of their dimensions (e.g. subjects, or voxels), and many algorithms work by simply concatenating the matrices along their shared dimension. 

\begin{figure}
    \centering
    \includegraphics[width=0.6\linewidth]{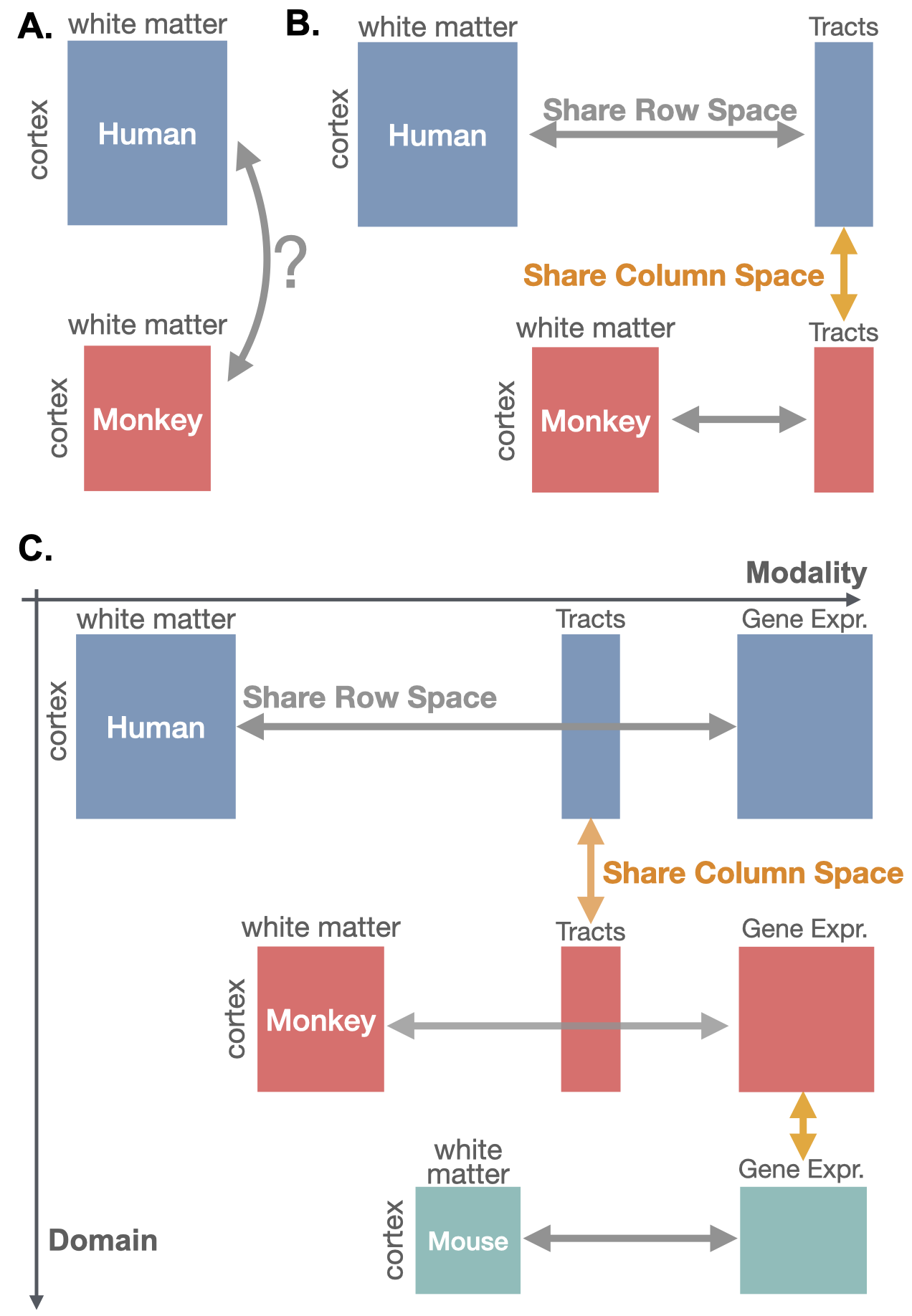}
    \caption{\textbf{PathFinder concept.} \textbf{(A)} Cortex $\times$ white matter matrices of two species that have incompatible dimensions. \textbf{(B)} Introducing subgroups of matrices that do have a dimension in common links all matrices, acting as a "path". \textbf{(C)} This can generalise to multiple domains and modalities with some modalities potentially missing for some domains. }
    \label{fig:concept}
\end{figure}

In some applications however, it is not always possible to find one common dimension across datasets. For example, comparing resting-state fMRI data across two different species cannot be achieved by concatenating the data. Neither the spatial nor the temporal dimensions are compatible. In multi-subject ICA this issue is resolved by analysing the subjects in standard space through nonlinear registration. This is not possible across species. Another example is missing data. For instance, we may want to jointly analyse connectivity and gene expression across humans, monkeys, and mice, but we may not have access to gene expression in monkeys. Can we still co-analyse all the data together?

We propose that joint outer-product decomposition can be generalised to situations where there is no single common dimension or where there is missing data. Suppose we have a collection of data matrices that we want to jointly analyse. Our method, PathFinder, has two requirements: (i) that pairs or sub-groups of the data matrices do have a dimension in common (either row- or column-dimension), and (ii) that there are one or more paths that link across all the data matrices (see Figure~\ref{fig:concept}). With this setup, a simple iterative scheme can be used to estimate a joint decomposition. Missing data matrices can be predicted from an outer-product of their corresponding components. 

In this paper we introduce the basic principles and algorithms behind the PathFinder method. We show that it can be combined with ICA to enable more interpretable latent components. We also show that several methods, such as linked-ICA or dual regression, can be cast under the PathFinder umbrella. Finally, we apply PathFinder in two real datasets: one where we predict the stimulus in a retinotopic task, and one where we analyse cortico-subcortical connectivity loops in diffusion MRI tractography data.

\section{Methods}

\subsection{Problem Formulation}\label{formulation}

We denote each dataset as a matrix $\mathbf{X}_{dm}$ where $d \in \{1,\dots,D\}$ indexes the domain (e.g. different species) and $m \in \{1,\dots,M\}$ indexes the modality (e.g. FMRI, Gene Expression, etc.). Matrix $\mathbf{X}_{dm}$ has dimensions $r_d \times c_m$, meaning all matrices within domain $d$ share $r_d$ rows, and all matrices within modality $m$ share $c_m$ columns.

We don't assume that all domain-modality combinations necessarily exist. Thus, we define $\mathcal{M} = \{(d,m) \mid \mathbf{X}_{dm} \text{ exists}\}$ as the set of available matrix indices. 

Our goal is to find low-rank decompositions of the form $\mathbf{X}_{dm} = \mathbf{A}_d \mathbf{S}_m^T$ for all available matrices. For a rank $r$ decomposition (i.e., $r$ components), we have $\mathbf{A}_d \in \mathbb{R}^{r_d \times r}$ and $\mathbf{S}_m \in \mathbb{R}^{c_m \times r}$. This formulation enforces that all matrices from domain $d$ share the same left factors $\mathbf{A}_d$, and all matrices from modality $m$ share the same right factors $\mathbf{S}_m$. The factors thus capture domain-specific and modality-specific subspaces. 

We minimise the reconstruction error with L2 regularisation. This avoids overfitting and is beneficial for prediction of unseen data (see Results section). However, it does not avoid the problem that solutions are not unique. This issue is mitigated by using independent component analysis post-hoc (see section \ref{ICA}).
\begin{equation}\label{loss}
\text{Loss} = \sum_{(d,m)\in\mathcal{M}} \left\lVert \mathbf{X}_{dm}-\mathbf{A}_d\mathbf{S}_m^T \right\rVert^2_{\mathrm{F}} + \alpha \sum_{d=1}^D \left\lVert \mathbf{A}_d \right\rVert^{2}_\mathrm{F} + \alpha \sum_{m=1}^M \left\lVert \mathbf{S}_m \right\rVert^2_\mathrm{F}
\end{equation}
where $||.||_F$ is the Frobenius norm and $\alpha$ controls the regularisation strength. 

This joint optimisation allows us to find relationships between incompatible matrices. For example, even if two matrices $\mathbf{X}_{d_1m_1}$ and $\mathbf{X}_{d_2m_2}$ have neither domain nor modality in common, their left and right factors $(\mathbf{A}_{d_1},\mathbf{A}_{d_2},\mathbf{S}_{m_1},\mathbf{S}_{m_2})$ are related through intermediate matrices $\mathbf{X}_{d_1m_2}$ or $\mathbf{X}_{d_2m_1}$ if they exist.

\subsection{Optimisation via Alternating Least Squares} \label{optimisation}

We use a simple alternating least squares algorithm to minimise the loss function. The algorithm alternates between updating all $\mathbf{A}_d$'s with $\mathbf{S}_m$'s fixed, and updating all $\mathbf{S}_m$'s with $\mathbf{A}_d$'s fixed. Each update has a closed-form solution.

\paragraph{Updating left factors} For a given domain $d$, we collect all matrices $\mathbf{X}_{dm}$ where $(d,m) \in \mathcal{M}$ and concatenate them horizontally into $\tilde{\mathbf{X}}_d = [\mathbf{X}_{dm_1}, \mathbf{X}_{dm_2}, \dots]$. We similarly concatenate the corresponding right factors vertically: $\tilde{\mathbf{S}} = [\mathbf{S}_{m_1}; \mathbf{S}_{m_2}; \dots]$. The left factor $\mathbf{A}_d$ is found using a ridge regression:

\begin{equation}
\mathbf{A}_d = \tilde{\mathbf{X}}_d \tilde{\mathbf{S}} \left(\tilde{\mathbf{S}}^T \tilde{\mathbf{S}} + \alpha\mathbf{I}\right)^{-1}
\end{equation}

\paragraph{Updating right factors} For a given modality $m$, we collect all matrices $\mathbf{X}_{dm}$ where $(d,m) \in \mathcal{M}$ and concatenate them vertically into $\hat{\mathbf{X}}_m = [\mathbf{X}_{d_1m}; \mathbf{X}_{d_2m}; \dots]$. We similarly concatenate the left factors: $\tilde{\mathbf{A}} = [\mathbf{A}_{d_1}; \mathbf{A}_{d_2}; \dots]$. The solution for the right factor $\mathbf{S}_m$ is given by:

\begin{equation}
\mathbf{S}_m = \hat{\mathbf{X}}_m^T \tilde{\mathbf{A}} \left(\tilde{\mathbf{A}}^T \tilde{\mathbf{A}} + \alpha\mathbf{I}\right)^{-1}
\end{equation}

Initialisation of $\mathbf{A}$ and $\mathbf{S}$ is left to the user. In our simulations we used random multivariate Gaussians. For real data, bespoke initialisations might be preferred. For example, joint singular value decomposition (see supplementary files) works well. Each subproblem is convex, so the loss function is monotonically non-increasing and converges. In practice, we run for a fixed number of iterations (typically 25 iterations) or until the relative change in loss falls below a threshold. 

This algorithm is summarised below. For large datasets, a batched version of this algorithm is described in detail in the supplementary files.

\begin{table}[h!]
\centering
\begin{tabular}{l l}
\hline
\multicolumn{2}{c}{\textbf{Algorithm: Alternating Least Square in PathFinder}} \\ \hline
  1: & Initialise $\mathbf{A_{\text{d}}}$ and $\mathbf{S_{\text{m}}}$ for $d=1,2,...D$ and $m=1,2,...M$\\
  2: & \textbf{while} not converged \textbf{do} \\
   & \hspace{2em} for $d=1,2,...D$ \\

  3a: & \hspace{2em}\hspace{2em}horizontally concatenate $\mathbf{X}_{\text{d,m}}$ as $\mathbf{\Tilde{X}}_{\text{d}}$ and $\mathbf{S}_{\text{m}}$ as $\mathbf{\Tilde{S}}$, for $m$ if $\mathbf{X}_{\text{dm}}$ exists\\
  3b: & \hspace{2em}\hspace{2em} set $\mathbf{A_{\text{d}}} \leftarrow \mathbf{\Tilde{X}_{\text{d}}}\mathbf{\Tilde{S}}(\mathbf{\Tilde{S}}^{T}\mathbf{\Tilde{S}} + \alpha\mathbf{I})^{-1}$ \\
     & \hspace{2em} for $m=1,2,...M$ \\
  4a: & \hspace{2em}\hspace{2em}vertically concatenate $\mathbf{X}_{\text{d,m}}$ as $\mathbf{\hat{X}}_{\text{m}}$ and $\mathbf{A}_{\text{d}}$ as $\mathbf{\Tilde{A}}$, for $d$ if $\mathbf{X}_{\text{dm}}$ exists\\
  4b:  & \hspace{2em}\hspace{2em} set $\mathbf{S_{\text{m}}} \leftarrow \mathbf{\hat{X}}_{\text{m}}^{T}\mathbf{\Tilde{A}}(\mathbf{\Tilde{A}}^{T} \mathbf{\Tilde{A}} + \alpha\mathbf{I})^{-1}$ \\

  6: & \textbf{end while} \\\hline
\end{tabular}
\caption{Pseudo code for the PathFinder alternating least squares algorithm.}
\label{tab:jointdecomp}
\end{table}

\subsection{ICA post-processing} \label{ICA}

The components of the loss function (\ref{loss}) are all invariant under a rotation $\mathbf{R}$ of the right and left factors, as:

\begin{equation}
\mathbf{A}_d\mathbf{S}_m^T=\mathbf{A}_d\mathbf{R}\mathbf{R}^T\mathbf{S}_m^T=(\mathbf{A}_d\mathbf{R})(\mathbf{S}_m\mathbf{R})^T,
\end{equation}

and similarly:

\begin{equation}
||\mathbf{A}_d||_F^2 = ||\mathbf{A}_d\mathbf{R}||_F^2, \text{  and  } ||\mathbf{S}_m||_F^2 = ||\mathbf{S}_m\mathbf{R}||_F^2
\end{equation}

So the algorithm only recovers the column spaces of $\mathbf{A}_d$ and $\mathbf{S}_m$. To resolve this ambiguity we can use independent component analysis (ICA) to find the rotation that maximises non-gaussianity, which tends to lead to more interpretable factors in practice. As we need to find a single rotation (unmixing) matrix, we can concatenate all the left-factors $\mathbf{A}_d$'s or all the right-factors $\mathbf{S}_m$'s vertically, and run ICA on that. One can also concatenate all of the left and right factors together to derive the unmixing matrix. The choice of either approach depends on the application, for example we may have prior reasons to expect some of the factors to be sparse or non-Gaussian (e.g., spatial maps for FMRI, temporal signals for event-driven data, etc.). 

\section{Results}
\subsection{Simulations - Face validity}
\begin{figure}[h]
    \centering
    \includegraphics[width=0.82\linewidth]{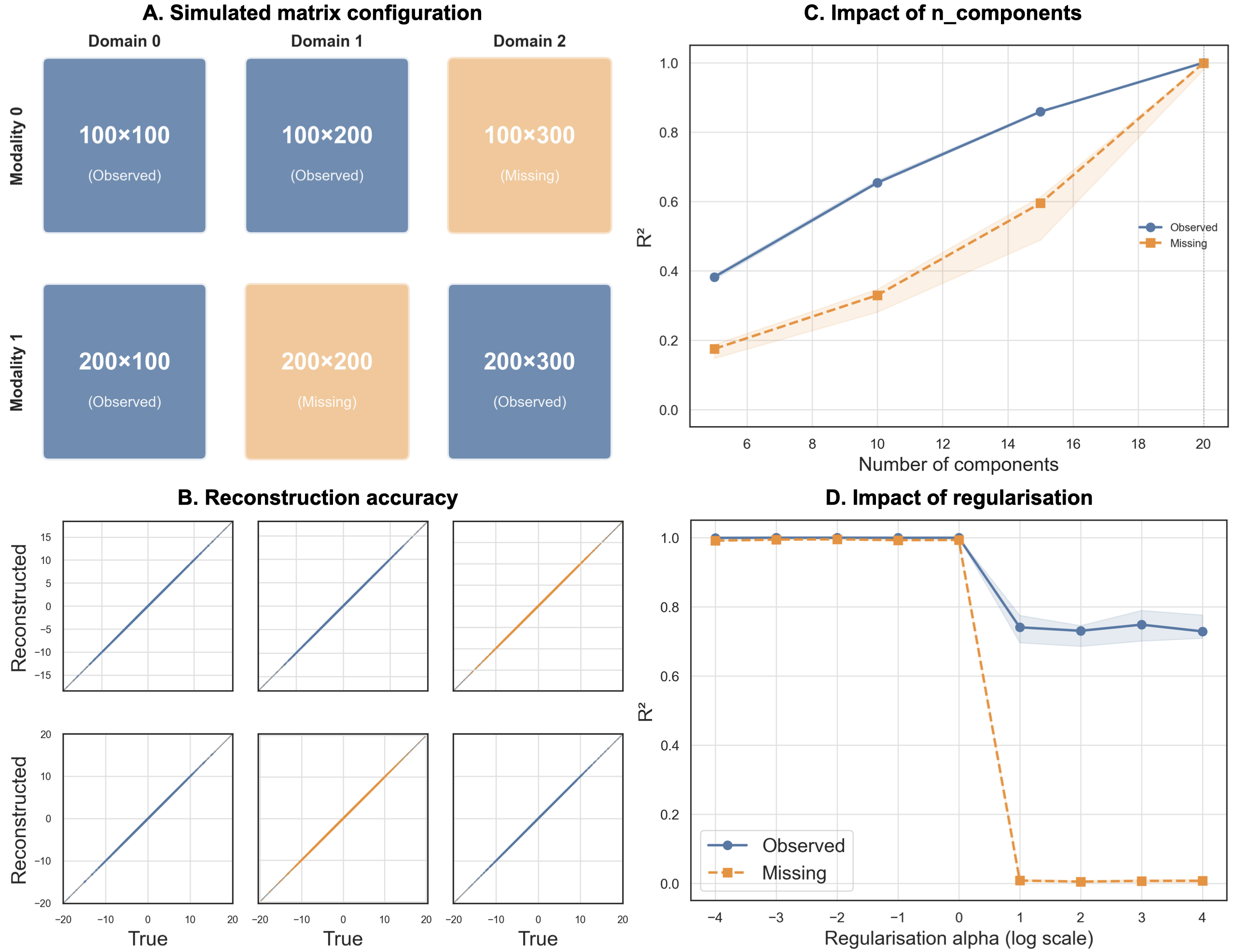}
    \caption{\textbf{Simulation results of a 2 x 3 grid of matrices spanning 2 modalities and 3 domains. }Four of the six matrices (blue) are provided to PathFinder for fitting, while two (orange) were marked as missing. \textbf{(A)} Illustration of matrix configuration. \textbf{(B)} Predicted values are plotted against ground truth at noise level $\sigma=0.1$ and $r=20$. All six matrices gave high reconstruction accuracy $R^2$. \textbf{(C)} The effect of rank $r$ (number of components) at noise level 0.01. Shaded regions indicate the interquartile range over 10 runs. \textbf{(D)} Impact of the regularisation parameter. }
    \label{fig:2}
\end{figure}

We first assess face validity on a simulated 2×3 grid of matrices spanning 2 modalities and 3 domains (Figure~\ref{fig:2}). Matrices on the same row share a modality factor (A matrix) while those on the same column share a domain factor (S matrix). The ground-truth rank was set to 20. To simulate all six matrices, we generated domain factors $\mathbf{A}_d $ and modality  factors $\mathbf{S}_m$ from random multivariate Gaussians, took their outer-product, and added Gaussian noise scaled relative to the signal norm. To test PathFinder's ability to impute missing data, we marked two matrices, $\mathbf{X}_{02}$ and $\mathbf{X}_{11}$, as "missing" and excluded them from model fitting. 

PathFinder accurately recovered both the observed and missing matrices (Figures~\ref{fig:2}B,C), with $R^2$ improving as the number of components increased and reaching its maximum at the true rank. The choice of regularisation parameter $\alpha$ is important. In this simulation, accuracy remained high for small values but declined as $\alpha$ increased. The unseen matrices degraded faster than the observed ones, which is expected when the problem is over-regularised (Figure~\ref{fig:2}D). In real application, the regularisation parameter can be set using cross-validation.

We next tested robustness to noise, benchmarking against Singular Value Decomposition (SVD) applied independently to each matrix (Figure~\ref{fig:svd}). This comparison was restricted to the observed data matrices, since independent SVDs cannot reconstruct unobserved data. PathFinder has higher $R^2$ than independent SVDs at higher noise levels ($\sigma > 0.5$). This is likely due to the effective denoising achieved by pooling information across modalities and domains compared with single-matrix decompositions.

\subsection{Simulations - Cyclic data organisation}


In this simulation we generate data that are organised in a cyclic structure. We will encounter a similar case in real data applications later where we model cortico-subcortical connectivity loops in diffusion MRI tractography.

\begin{figure}[h]
    \centering
    \includegraphics[width=0.95\linewidth]{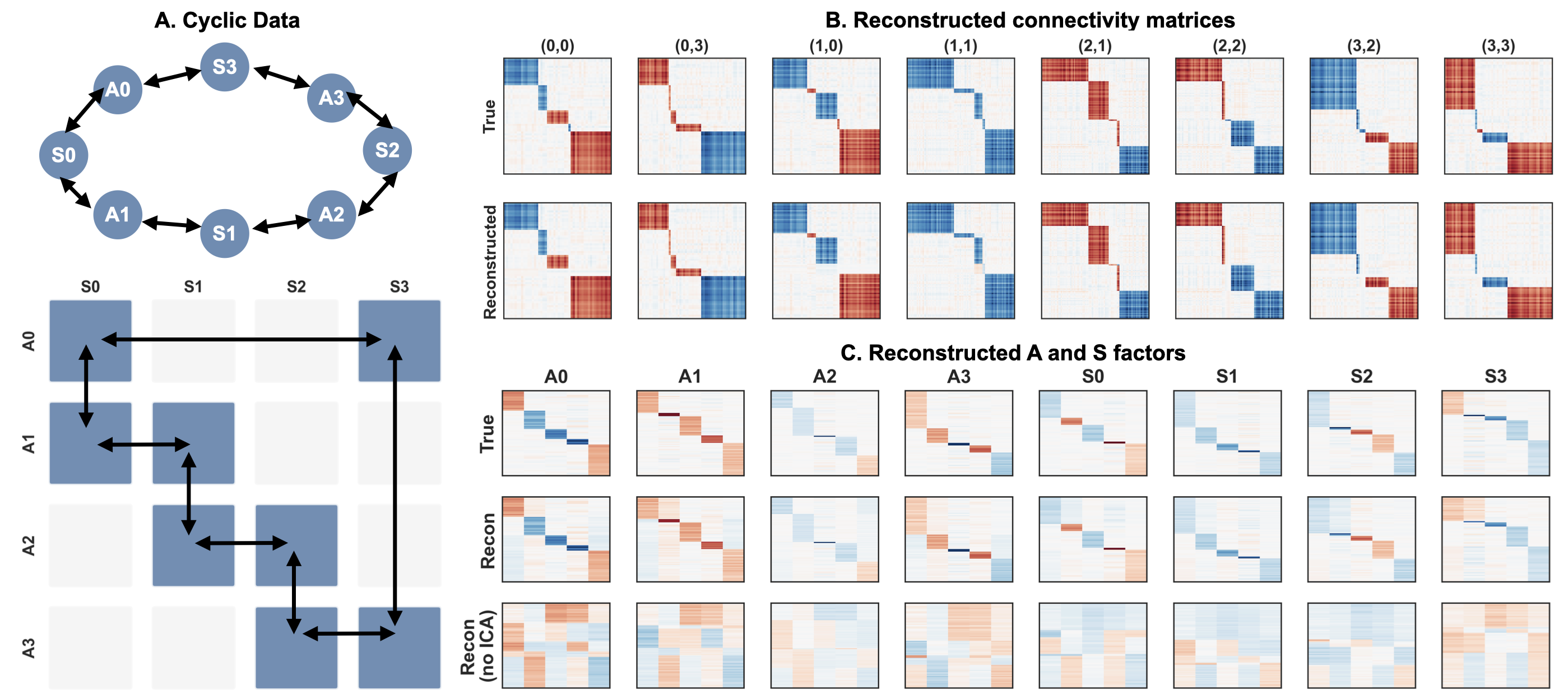}
    \caption{\textbf{Simulation results of cyclic data decomposition.} \textbf{(A)} Example of cyclic data and its PathFinder configuration. \textbf{(B)} True and predicted activity. \textbf{(C)} True and reconstructed factors (with or without ICA post-processing).}
    \label{fig:cyclic-simulation}
\end{figure}

What we mean by cyclic data is a dataset in which the factors (A and S matrices) form a closed loop (see Figure~\ref{fig:cyclic-simulation}A). Each factor is connected to two neighbours, and pairs of adjacent factors combine to form an outer-product model. Arranged as a table, this means that the data to be modelled can be arranged by alternatively concatenating pairs along rows and colums, where the last matrix shares its row-space with the initial matrix. 

Similar to our previous simulation, the data is recovered well  (Figure~\ref{fig:cyclic-simulation}B). However, the true factors are not (Figure~\ref{fig:cyclic-simulation}C, compare first and third row). Regularisation is not sufficient, as the factors are only uniquely determined up to a rotation. However, concatenating all the reconstructed factors and running ICA, we can recover the ground truth simulated factors (Figure~\ref{fig:cyclic-simulation}C, compare first and second row).

\subsection{Reconstructing the stimulus from brain data}

Next we turn to real data. Our first example uses a retinotopy task FMRI dataset from the Human Connectome Project acquired at 7T \citep{Benson2018}. Data was averaged over all subjects, and the time series from primary visual cortex (V1) extracted. The task comprises 6 runs of visual stimulation: clockwise/counter-clockwise rotating wedges, expanding/contracting rings, and drifting bars. 

\begin{figure}[h]
    \centering
    \includegraphics[width=0.7\linewidth]{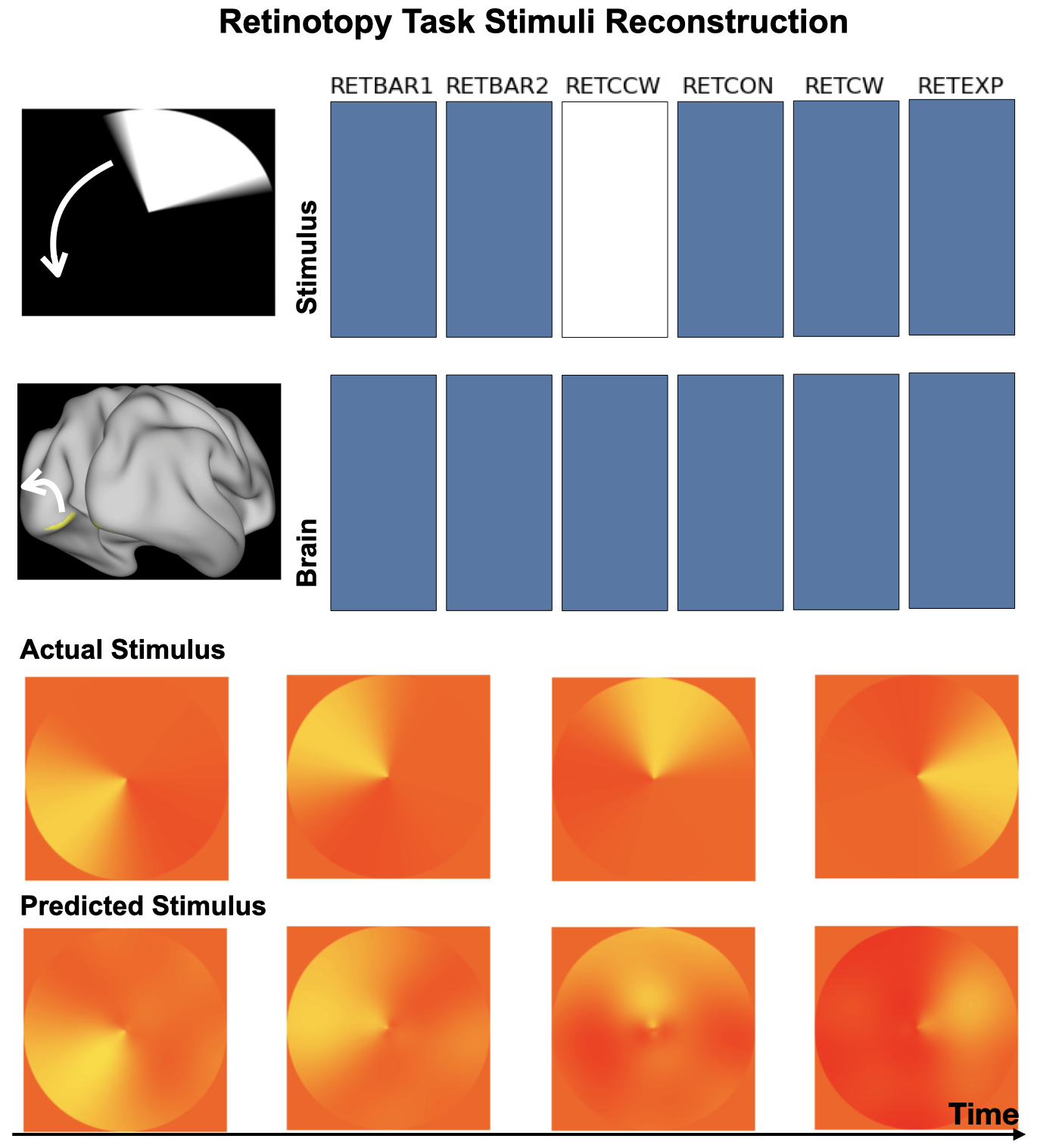}
    \caption{\textbf{Example of PathFinder reconstructing missing retinotopy task stimuli.} Brain grayordinates and visual field pixels served as domains. The six different retinotopy tasks served as modalities. the RETCCW stimulus (Counter-clockwise rotating wedge) was marked as missing and held out from PathFinder training. The predicted RETCCW matches the actual stimulus.}
    \label{fig:retinotopy}
\end{figure}

We used PathFinder to model both the stimuli and the FMRI time courses using an outer-product model. The data was arranged as Nxtime matrices where N represents pixels for the stimuli and cortical vertices for the FMRI data. We arranged these paired time series as a 2x6 grid (Figure~\ref{fig:retinotopy}), in which the two rows represent the brain voxel domain and the visual-field pixel domain, while the six columns index the 6 runs. This configuration assumes that the two "domains" share their model time courses (S matrices). To account for haemodynamic lag, the stimulus time course was convolved with a canonical double-gamma haemodynamic response function.  

We asked if PathFinder can recover one of the stimuli from the remaining stimuli and FMRI data. We held out the counter-clockwise rotating wedge stimulus, then fitted PathFinder to the remaining 11 datasets. We then predicted the held-out stimulus as the outer product of the stimulus row factor and the run's column factor. PathFinder was able to recover the held-out stimulus (Figure~\ref{fig:retinotopy}, lower panel). Interestingly, repeating the same procedure after excluding one hemisphere broke the prediction. In this case PathFinder was only able to partially recover the stimulus, i.e. in one hemifield (see supplementary movies).

\subsection{Equivalence with Dual-Regression}
We next show how PathFinder can be set up to perform a dual-regression analysis of resting state FMRI. 

Dual regression is a common method used to map resting-state brain maps in individual subjects while preserving one-to-one correspondence in the identities of the maps across subjects \citep{Nickerson2017}. It is a two-stage procedure that starts with a set of group-level spatial maps. This is typically from group ICA of concatenated time courses across subjects. In the first stage, each subject's 4D FMRI time series is regressed against the group maps to obtain subject-specific time courses. In the second stage, those time courses are regressed back against the subject's data to obtain subject-specific spatial maps. This yields matched spatial maps and time courses for every subject while modelling individual differences in the maps. 

\begin{figure}[h]
    \centering
    \includegraphics[width=\linewidth]{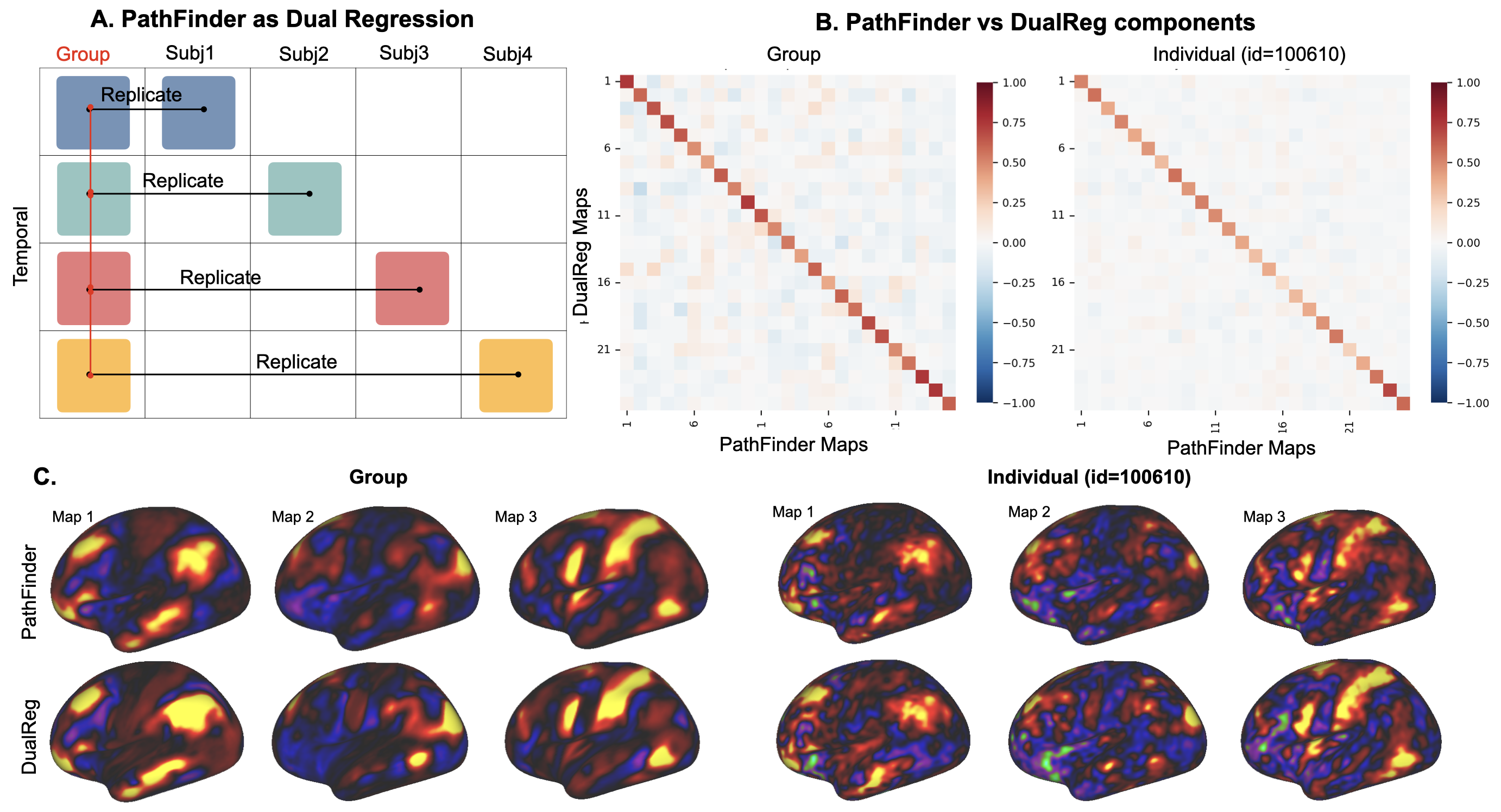}
    \caption{\textbf{PathFinder for Dual-Regression.} \textbf{(A)} Example matrix configuration for PathFinder as Dual-Regression \textbf{(B)} PathFinder components are similar to dual regression spatial maps, both at the level of the group and individuals. \textbf{(C)} Visualised PathFinder and dual regression maps for the group and an example subject.}
    \label{fig:dr}
\end{figure}

PathFinder can be set up to achieve the same goal in a single joint decomposition (Figure~\ref{fig:dr}A). The FMRI time series of all $N$ subjects are concatenated along the time axis to form a single concatenated group column-space, and each subject is additionally given its own column containing only that subject's data. In this  layout, the spatial factor $\mathbf{S}_0$ of the group column is constrained by every subject's data and therefore captures group-level spatial factors, while each  subject-column factor $\mathbf{S}_j$ captures that subject's individual spatial maps. The temporal factors $\mathbf{A}_i$ are the subject-specific time courses and are jointly estimated. PathFinder thus recovers group spatial maps, subject-specific spatial maps, and subject-specific time courses simultaneously. The recovered maps show high correspondence with those from traditional dual regression, both for the group and individual levels (Figure~\ref{fig:dr}B and 5C).

\subsection{Revealing cortico-basal loops with cyclic PathFinder}
\begin{figure}[h]
    \centering
    \includegraphics[width=\linewidth]{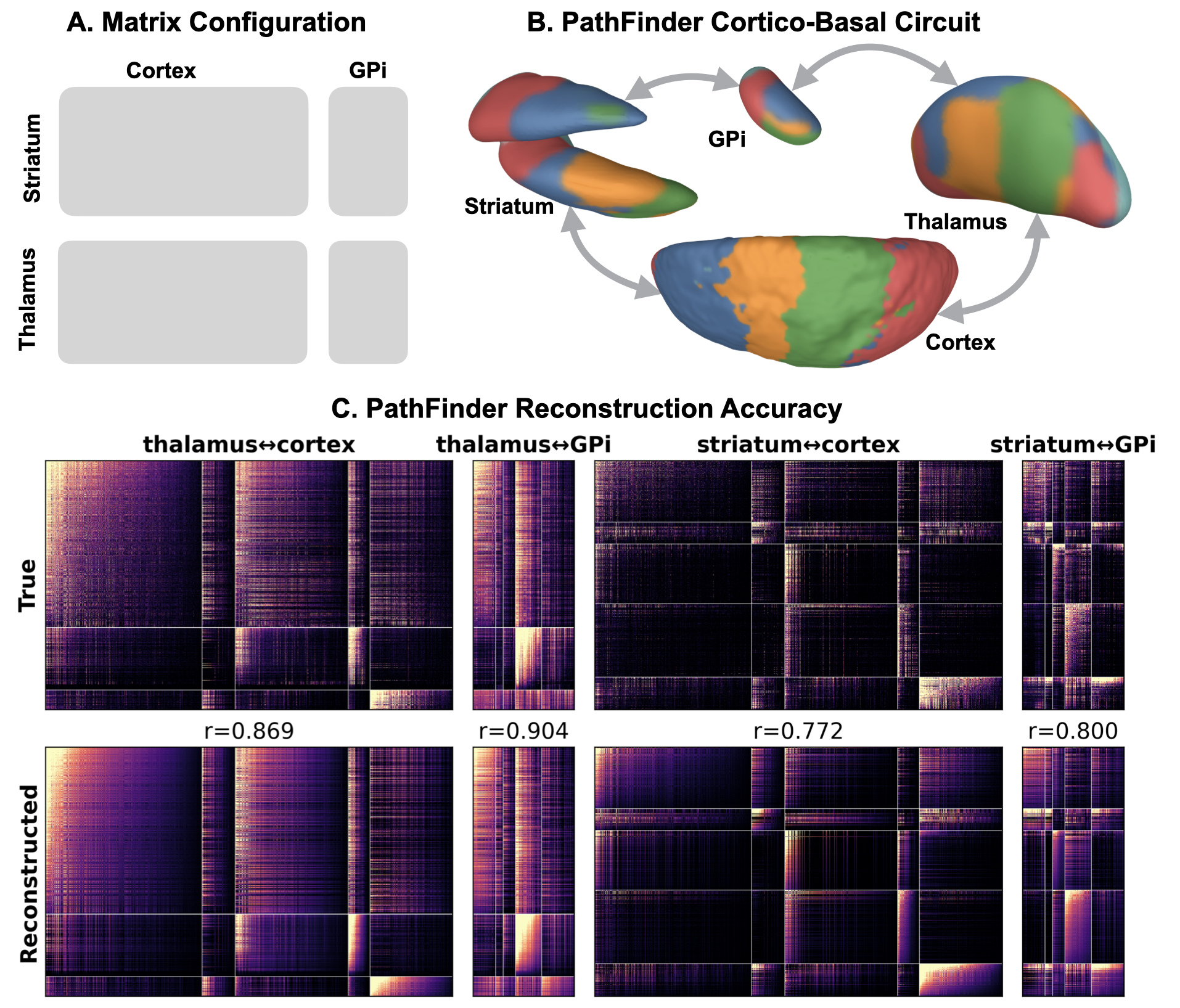}
    \caption{\textbf{PathFinder decomposition of the cortico-basal direct loop.} A: PathFinder configuration. B: running a winner-takes-all across components reveals functional circuits across the loop. C: connectivity matrices and their low-rank reconstructed counterparts. }
    \label{fig:thalamo}
\end{figure}

As a real-data example of the cyclic decomposition mentioned above, we applied PathFinder followed by post-hoc ICA to reveal spatial organisation of cortico-basal connectivity loops, a well-established set of circuits in the brain. Using 3T diffusion MRI data from 10 Human Connectome Project subjects \citep{VanEssen2012,Sotiropoulos2013}, we estimated a voxelwise connectivity matrix for each pair of connected regions in the loop using FSL's BedpostX and probabilistic tractography \citep{Behrens2007,Jbabdi2012}, and averaged the results across subjects. Note that we only included the regions in the direct pathway of the basal ganglia, including the cortex, thalamus, striatum, and Globus Pallidus Internus (GPi). The connectivity matrices were tiled into a 2×2 grid in which rows and columns index the successive regions forming the loop. Each matrix represents the structural connectivity between a row and a column region (Figure~\ref{fig:thalamo}A). The PathFinder-estimated factors provide a latent representation of the connectivity in terms of its spatial arrangement with each region  (Figure~\ref{fig:thalamo}B). 

To interpret the factors anatomically, we assigned every voxel to its strongest-loading component (winner-takes-all). Because components are shared across regions, voxels from different regions sharing a component belong to the same functional circuit. This recovered the three canonical basal ganglia pathways, motor, associative, and limbic (Figure~\ref{fig:thalamo}C).

\section{Discussion}

PathFinder is a simple but general framework for joint matrix decomposition. We have seen how several other approaches, such as joint ICA or dual regression fall under this umbrella. But the real added benefit of PathFinder is when combining data from different domains, with potentially missing subsets. This is not unusual when analysing data across species. For example, it is almost impossible to find functional MRI data of chimpanzees, but relatively straightforward to acquire diffusion MRI in any species. Some datasets can serve as an anchor, then we can use this framework to find joint decompositions, leading to the possibility of analyses in a common subspace. We can use this subspace to compare different species, or to make predictions and test them against ground truth data to understand species commonalities and differences. The algorithm is very simple: iterate through the A and S components and estimate using data concatenation. To enable analysis of large matrices, we also introduce and implemented a batched version, allowing whole brain fitting. Performance is extensively tested in supplementary material. One clear limitation is the ill-posed nature of outer-product decompositions. This is not resolved by our L2-regularisation, which merely mitigates overfitting. In order for the components to be unique and more interpretable, additional constraints must be added. We showed in the diffusion tractography experiment that ICA can be applied post-hoc to rotate the components. In simulations, this allowed us to recover the ground truth components. In real data, this leads to an anatomically plausible parcellation of all the elements of the cortico-basal direct loop. Other constraints can easily be incorporated. For instance, one could use non-negative matrix factorisation to aid identifiability, assuming all the datasets have positive entries. In the remainder of this discussion we propose several other future extensions to PathFinder. 

One way to obtain a unique decomposition (up to sign flip) is via singular value decomposition (SVD). An extension to SVD is joint SVD (see \citep{Congedo2011}, and algorithm in the supplementary material), which can be adopted to be used in the PathFinder setting, i.e. with potentially missing datasets. Joint SVD resolves uniqueness by imposing orthogonality and isolating the weights (singular values) from the factors. This last property has the additional advantage that each data matrix can have its own set of weights. For example component 1 can be dominant in one modality but less dominant in another. The downside of joint SVD is that we cannot predict missing modalities as we don't know the weights for the missing data. However, it can serve as a method for initialising the outer-product version of PathFinder.

\begin{figure}
    \centering
    \includegraphics[width=0.8\linewidth]{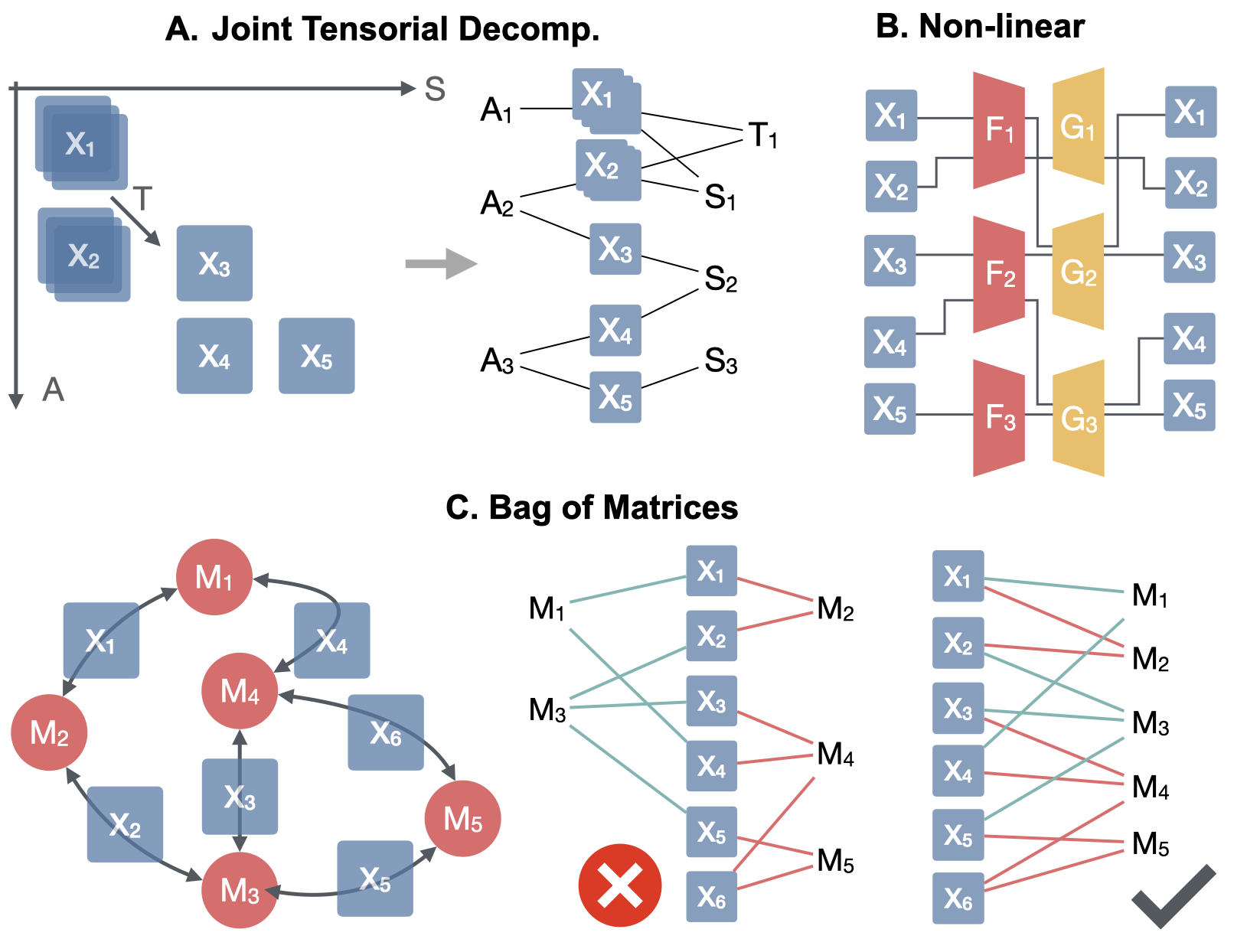}
    \caption{\textbf{PathFinder Extensions.} \textbf{(A)} PathFinder can accommodate not only 2D matrices but also higher-order tensors. \textbf{(B)} An example of non-linear PathFinder using the encoder-decoder architecture. \textbf{(C)} PathFinder can drop the bipartite formulation, i.e., arranging the factors on one side instead of on left and right. This expands the applications for PathFinder, allowing it to run natively on biological networks.}
    \label{fig:extension}
\end{figure}

Another extension is tensor decomposition \citep{Beckmann2005tensorial}, where some of the datasets come in the form of high order tensors with more than two dimensions (Figure~\ref{fig:extension}A). 

Low-rank outer-product decompositions can be re-interpreted as linear low-dimensional embeddings. To see this, consider a decomposition $\mathbf{X}\approx\mathbf{A}\mathbf{S}^T$. If $\mathbf{X}$ is a tall matrix, this can be re-written as $\mathbf{X}\mathbf{S}^\dagger\approx\mathbf{A}$ ($\mathbf{S}^\dagger$ being the pseudo-inverse of $\mathbf{S}$). For each row vector $\mathbf{x}^r_i$ of $\mathbf{X}$ we thus have the linear embedding (encoding) $\mathbf{x}^r_i\mathbf{S}^\dagger=\mathbf{a}^r_i$. The reconstruction is given by $\tilde{\mathbf{x}}^r_i=\mathbf{a}^r_i\mathbf{S}^T$ (decoding). Thus a natural generalisation is to use non-linear encoders and decoders (a.k.a. auto-encoders, see Figure~\ref{fig:extension}B).

A major limitation that we have not touched upon yet is that PathFinder as it stands requires that the data matrices can be arranged as a 2D matrix (Figure~\ref{fig:concept}). With this type of configuration, there is a set of A matrices and set of S matrices, and these two sets are distinct, i.e. an A matrix for some dataset cannot be an S matrix for another dataset (Figure~\ref{fig:extension}C, middle panel). This limits applicability. For instance, consider the cortico-basal loop example. We were lucky with the direct loop as it only has 4 ROIs that can neatly be arranged into a 2D table with no overlap between the A and S sets. But if we were to study the indirect loop which has 5 regions, then the A matrix for the 5th region must also coincide with the A matrix of the 1st, which does not fit with the 2D tabular arrangement of the data. A more flexible framework is shown in Figure~\ref{fig:extension}C (right panel). Here the components are arranged into a bag of matrices $\left\{\mathbf{M}_k\right\}$, from which the data are to be decomposed. To decompose a given matrix, we choose two factors, one for the left components and one for the right components, both from the same bag of matrices. Thus any given $\mathbf{M}_k$ matrix can play the role of a left or right matrix. 

Finally, the implementation of PathFinder have been released \href{https://doi.org/10.5281/zenodo.21383855}{https://doi.org/10.5281/zenodo.21383855}. Data and supplementary files associated with the experiments can be found under \href{https://osf.io/jz8vf/overview}{https://osf.io/jz8vf/overview}.

\newpage

\bibliographystyle{unsrtnat}
\bibliography{references}  







\setcounter{equation}{0}
\setcounter{figure}{0}
\setcounter{table}{0}

\renewcommand{\theequation}{S\arabic{equation}}
\renewcommand{\thefigure}{S\arabic{figure}}
\renewcommand{\thetable}{S\arabic{table}}

\clearpage
\newpage

\begin{center}

\textbf{\large Supplemental Materials}
\end{center}

\section{Mini-batch implementation of Joint Decomposition}
On large datasets such as MRI images, individual matrices $\mathbf{X}_{\text{dm}}$ can have thousands of rows and columns, making the concatenated matrix $\Tilde{\mathbf{X}}_d$ or $\hat{\mathbf{X}}_m$ prohibitively large for memory. We address this challenge by implementing an algorithm to process data in mini-batches instead of solving the full least squares at once during each iteration. 

Consider updating $\mathbf{A}_d$ with $\mathbf{S}_m$ fixed. We partition each matrix $\mathbf{X}_{dm}$ along its columns into $B$ batches: $\mathbf{X}_{dm}=[\mathbf{x}^{1}_{\text{dm}}, \mathbf{x}^{2}_{\text{dm}},...\mathbf{x}^{B(m)}_{\text{dm}}]$, where $B(m)$ is the number of mini-batches for modality $m$. The corresponding rows of $\mathbf{S}_m$ are partitioned accordingly: $\mathbf{S}_{\text{m}}=[\mathbf{s}^{1}_{m}; \mathbf{s}^{2}_{m};...;\mathbf{s}^{B(m)}_{m}]$, where each $\mathbf{S}_{m}^{b}$ contains the rows corresponding to batch $b$. 

With this partitioning, the loss function becomes:

\begin{equation}
\sum_{(d,m)\in\mathcal{M}}\sum_{b=1}^{B(m)} \left\lVert \mathbf{x^{\text{b}}_{\text{d,m}}}-\mathbf{A_{\text{d}}}\mathbf{s^{\text{b}}_{\text{m}}}^{T}\right\rVert^2_{\mathrm{F}} + \alpha \sum_{d=1}^D \left\lVert \mathbf{A_{\text{d}}} \right\rVert^{2}_{\mathrm{F}} + \alpha \sum_{m=1}^M \sum_{b=1}^{B(m)}\left\lVert \mathbf{s^{\text{b}}_{\text{m}}} \right\rVert^2_2
\end{equation}

Setting the derivative of the loss function with respect to $\mathbf{A}_d$ to zero, we find the optimal $\mathbf{A}_{d}$ as:

\begin{equation}
\mathbf{A}_{\text{d}}=(\sum_{m\in\mathcal{M}_d}\sum_{b=1}^{B(m)}\mathbf{x^{\text{b}}_{\text{d,m}}}\mathbf{s^{\text{b}}_{\text{m}}})(\sum_{m\in\mathcal{M}_d}\sum_{b=1}^{B(m)}\mathbf{s^{\text{b}}_{\text{m}}}^{T}\mathbf{s^{\text{b}}_{\text{m}}}+\alpha\mathbf{I})^{-1}
\end{equation}
where $\mathcal{M}_d$ is the set of modalities for domain $d$.

The key advantage of this mini-batch approach is that both the numerator and denominator are sums over batches. We can accumulate the cross product $\sum_{b}(\mathbf{x^{\text{b}}_{\text{d,m}}}\mathbf{s^{\text{b}}_{\text{m}}})$ and the gram matrix $\sum_{b}\mathbf{s^{\text{b}}_{\text{m}}}^{T}\mathbf{s^{\text{b}}_{\text{m}}}$ incrementally, processing one batch at a time without requiring storage of the full concatenated matrices $\Tilde{\mathbf{S}}$ or $\Tilde{\mathbf{X}}_{d}$ in memory.

The update for $\mathbf{S}_{m}$ follows symmetrically, partitioning matrices along their rows instead of columns. For each modality $m$, we partition $\mathbf{X}_{dm}$ into row batches and accumulate $\sum_{b}\mathbf{a^{\text{b}}_{\text{d}}}^{T}\mathbf{a^{\text{b}}_{\text{d}}}$ and $\sum_{b}\mathbf{x^{\text{b}}_{\text{d,m}}}^{T}\mathbf{a^{\text{b}}_{\text{d}}}$ to compute $\mathbf{S}_m$:

\begin{equation}
\mathbf{S}_{\text{m}}=(\sum_{d\in\mathcal{D}_m}\sum_{b=1}^{B(d)}\mathbf{x^{\text{b}}_{\text{d,m}}}^{T}\mathbf{a^{\text{b}}_{\text{d}}})(\sum_{d\in\mathcal{D}_m}\sum_{b=1}^{B(d)}\mathbf{a^{\text{b}}_{\text{d}}}^{T}\mathbf{a^{\text{b}}_{\text{d}}}+\alpha\mathbf{I})^{-1}
\end{equation}
where $\mathcal{D}_m$ is the set of domains for modality $m$, and $B(d)$ the number of batches.

The pseudo-code for mini-batch update is summarised in Table~\ref{tab:jointdecomp-minibatch}:

\begin{table}[h!]
\centering
\begin{tabular}{l l}
\hline
\multicolumn{2}{c}{\textbf{Algorithm: Minibatch Alternate Least Square in PATHFINDER}} \\ \hline
  1: & Initialise $\mathbf{A_{\text{d}}}$ and $\mathbf{S_{\text{m}}}$ for $d=1,2,...D$ and $m=1,2,...M$\\
  2: & \textbf{while} not converged \textbf{do} \\
   & \hspace{2em} for $d=1,2,...D$ \\

  3a: & \hspace{2em}\hspace{2em} initialise accumulators for updating $\mathbf{A}_{\text{d}}$, $\mathbf{C}_{\text{num}} \leftarrow 0$, $\mathbf{C}_{\text{denom}} \leftarrow 0$ \\
  3b:  & \hspace{2em}\hspace{2em} for $m\in\mathcal{M}_d$, $b\in\{1,2,...B(m)\}$\\
   & \hspace{2em}\hspace{2em}\hspace{2em} set  $\mathbf{C}_{\text{num}} \leftarrow \mathbf{C}_{\text{num}} + \mathbf{x^{\text{b}}_{\text{d,m}}}\mathbf{s^{\text{b}}_{\text{m}}} $ \\
   & \hspace{2em}\hspace{2em}\hspace{2em} set  $\mathbf{C}_{\text{denom}} \leftarrow \mathbf{C}_{\text{denom}} + \mathbf{s^{\text{b}}_{\text{m}}}^{T}\mathbf{s^{\text{b}}_{\text{m}}} $ \\
   & \hspace{2em}\hspace{2em} set $\mathbf{A_{\text{d}}} \leftarrow \mathbf{C}_{\text{num}}(\mathbf{C}_{\text{denom}} + \alpha\mathbf{I})^{-1}$ \\
     & \hspace{2em} for $m=1,2,...M$ \\

  4a: & \hspace{2em}\hspace{2em} initialise accumulators for updating $\mathbf{S}_{\text{m}}$, $\mathbf{C}_{\text{num}} \leftarrow 0$, $\mathbf{C}_{\text{denom}} \leftarrow 0$ \\
  4b:  & \hspace{2em}\hspace{2em} for $d\in\mathcal{D}_m$, $b\in\{1,2,...B(d)\}$\\
   & \hspace{2em}\hspace{2em}\hspace{2em} set  $\mathbf{C}_{\text{num}} \leftarrow \mathbf{C}_{\text{num}} + \mathbf{x^{\text{b}}_{\text{d,m}}}^{T}\mathbf{a^{\text{b}}_{\text{d}}} $ \\
   & \hspace{2em}\hspace{2em}\hspace{2em} set  $\mathbf{C}_{\text{denom}} \leftarrow \mathbf{C}_{\text{denom}} + \mathbf{a^{\text{b}}_{\text{d}}}^{T}\mathbf{a^{\text{b}}_{\text{d}}} $ \\
   & \hspace{2em}\hspace{2em} set  $\mathbf{S_{\text{m}}} \leftarrow \mathbf{C}_{\text{num}}(\mathbf{C}_{\text{denom}} + \alpha\mathbf{I})^{-1}$ \\
  6: & \textbf{end while} \\\hline
\end{tabular}
\caption{Pseudo code of Minibatch ALS algorithm to solve PATHFINDER.}
\label{tab:jointdecomp-minibatch}
\end{table}

\subsection{Implementation of Joint Singular Value Decomposition}
Joint Singular Value Decomposition (SVD) offers an alternative approach to joint matrix decomposition with orthogonality constraints, which can improve numerical stability and interpretability or be used for initialisation. For the same set of matrices $\mathbf{X}_{dm}$, we aim to find decompositions for every pair $(d,m)\in\mathcal{M}$:

\begin{equation}
\mathbf{X}_{d,m} = \mathbf{U}_d \mathbf{D}_{d,m} \mathbf{V}_m^T
\end{equation}

where $\mathbf{U}_d$ and $\mathbf{V}_m$ are orthogonal matrices of singular vectors and $\mathbf{D}_{dm}$ the diagonal matrix of singular values. Unlike the simple outer-product joint decomposition where factors are shared exactly, here each matrix has its own scaling through $\mathbf{D}_{dm}$ while sharing the orthogonal bases across domains and modalities.

Similarly, the algorithm to solve joint SVD alternates between updating $\mathbf{U}_d$, $\mathbf{V}_m$ and $\mathbf{D}_{dm}$ with orthogonality constraints. 

\paragraph{Updating $\mathbf{D}_{dm}$} given fixed $\mathbf{U}_d$ and $\mathbf{V}_m$, the optimal diagonal matrix is:

\begin{equation}
(\mathbf{D}_{d,m})_{nn} = (\mathbf{u}_d^{(n)})^T\mathbf{X}_{d,m}\mathbf{v}_m^{(n)}
\end{equation}
where $(\mathbf{D}_{dm})_{nn}$ is the n-th diagonal element of $\mathbf{D}_{dm}$.
This operation simply projects the data onto the current orthonormal bases and extracts the diagonal.

\paragraph{Updating $\mathbf{U}_{d}$} For a given $n$-th component of $\mathbf{U}_d$, we have $\mathbf{X}_{dm}\mathbf{v}_{m}^{(n)} \approx \mathbf{u}_{d}^{(n)} \cdot d_{d,m}^{(n)}$, where $\mathbf{v}_{m}^{(n)}$ is the $n$-th column of $\mathbf{V}_{m}$. Stacking all these equations, we have


\begin{equation}
[\mathbf{X}_{d,m_1}\mathbf{v}_{m_1}^{(n)},\mathbf{X}_{d,m_2}\mathbf{v}_{m_2}^{(n)} , \dots] \approx \mathbf{u}_d^{(n)}[d_{d,m_1}^{(n)}, d_{d,m_2}^{(n)},\dots]
\end{equation}

For each component $n\in\{1,...,r\}$, we define a matrix by stacking these projections $\mathbf{X}_{d,m}\mathbf{v}_{m}^{(n)}$ for all modalities $m\in\mathcal{M}_d$ at domain $d$:
\begin{equation}
\mathbf{M}^{(n)}_d(\mathbf{V}) = [\mathbf{X}_{d,m_1}\mathbf{v}_{m_1}^{(n)}, \mathbf{X}_{d,m_2}\mathbf{v}_{m_2}^{(n)}, \dots]  \approx \mathbf{u}_d^{(n)}[d_{d,m_1}^{(n)}, d_{d,m_2}^{(n)},\dots]
\end{equation}
$\mathbf{u}_d^{(n)}$, the $n-$th column of $\mathbf{U}_d$, is solved as the left leading singular vector of $\mathbf{M}_{d}^{(n)}(\mathbf{V})$. After computing all $r$ columns for $\mathbf{U}_d$, we orthogonalise $\mathbf{U}_d$ using a QR decomposition.

\paragraph{Updating $\mathbf{V}_{m}$} The update for $\mathbf{V}_m$ is symmetric. For each component $n$, we construct
\begin{equation}
\mathbf{M}^{(n)}_m(\mathbf{U}) = [\mathbf{X}_{d_1,m}^T\mathbf{u}_{d_1}^{(n)}, \mathbf{X}_{d_2,m}^T\mathbf{u}_{d_2}^{(n)}, \dots]
\end{equation}
where $\mathbf{u}_d^{(n)}$ is the $n$-th column of $\mathbf{U}_d$. Each column of $\mathbf{V}_m$ is solved as the leading left singular vector of the corresponding projection matrix, e.g., $\mathbf{M}_{m}^{(n)}(\mathbf{U})$. $\mathbf{V}_m$ is subsequently orthogonalised using the QR orthogonalisation.

\begin{table}[h!]
\centering
\begin{tabular}{l l}
\hline
\multicolumn{2}{c}{\textbf{Algorithm: Joint SVD}} \\ \hline
  1: & Initialise $\mathbf{U_{\text{d}}}$, $\mathbf{D}_{d,m}$, and $\mathbf{V_{\text{m}}}$ for $d=1,2,...D$ and $m=1,2,...M$\\
   & \textbf{while} not converged \textbf{do} \\
  2: & \hspace{2em} for $(d, m)\in \mathcal{M}$ \\
     & \hspace{2em}\hspace{2em} set $\mathbf{D}_{d,m} \leftarrow \mathbf{U}_d^{T}\mathbf{X}_{d,m}\mathbf{V}_m$ \\
  3: & \hspace{2em} for each domain $d=1,2,...D$ \\
     & \hspace{2em}\hspace{2em} for each component $n=1,2,...r$\\
  3a:   & \hspace{2em}\hspace{2em}\hspace{2em} construct matrix  $\mathbf{M}^{(n)}_d(\mathbf{V}) \leftarrow [\mathbf{X}_{d,m_1}\mathbf{v}_{m_1}^{(n)}, \mathbf{X}_{d,m_2}\mathbf{v}_{m_2}^{(n)}, \dots]$\\
  3b:   & \hspace{2em}\hspace{2em}\hspace{2em} set $\mathbf{U}_{d}^{(n)}\leftarrow \text{left leading singular vector of } \mathbf{M}_{d}^{(n)}(\mathbf{V})$\\
  3c: & \hspace{2em}\hspace{2em}\hspace{2em} set $\mathbf{U}_d \leftarrow \text{orthogonalise}([\mathbf{U}_{d}^{(1)}, \mathbf{U}_{d}^{(2)},.., \mathbf{U}_{d}^{(r)}])$ \\
 4: & \hspace{2em} for each modality $m=1,2,...M$ \\
     & \hspace{2em}\hspace{2em} for each component $n=1,2,...r$\\
  4a:   & \hspace{2em}\hspace{2em}\hspace{2em} construct matrix  $\mathbf{M}^{(n)}_m(\mathbf{U}) \leftarrow [\mathbf{X}_{d_1,m}^T\mathbf{u}_{d_1}^{(n)}, \mathbf{X}_{d_2,m}^T\mathbf{u}_{d_2}^{(n)}, \dots]$\\
  4b:   & \hspace{2em}\hspace{2em}\hspace{2em} set $\mathbf{V}_{m}^{(n)}\leftarrow \text{left leading singular vector of } \mathbf{M}_{m}^{(n)}(\mathbf{U})$\\
  4c: & \hspace{2em}\hspace{2em}\hspace{2em} set $\mathbf{V}_m \leftarrow \text{orthogonalise}([\mathbf{V}_{m}^{(1)}, \mathbf{V}_{m}^{(2)},.., \mathbf{V}_{m}^{(r)}])$ \\
   & \textbf{end while} \\\hline
\end{tabular}
\caption{Pseudo code of the algorithm to solve joint SVD.}
\label{tab:jointsvd}
\end{table}

\subsection{Mini-batch Implementation of JointSVD}
For Joint SVD, mini-batching takes a different form. When updating $\mathbf{U}_d$ for component $n$, constructing the full $\mathbf{M}_d^{(n)}(\mathbf{V})$ requires holding every projection $\mathbf{X}_{d,m}\mathbf{v}_m^{(n)}$ in memory, making it impossible to calculate the left singular vector at once. We therefore seek to obtain its leading left singular vector via power iteration on $\mathbf{M}_d^{(n)}(\mathbf{V})\mathbf{M}_d^{(n)}(\mathbf{V})^T$. 

 \paragraph{Power iteration.} Let $\mathbf{M}=\mathbf{M}_d^{(n)}(\mathbf{V})$ and  write its SVD as $\mathbf{M}=\sum_i \sigma_i \mathbf{p}_i \mathbf{q}_i^T$, with  $\sigma_1>\sigma_2\geq\dots$. Then $\mathbf{M}\mathbf{M}^T=\sum_i\sigma_i^2\,\mathbf{p}_i\mathbf{p}_i^T$, and starting from any vector $\mathbf{u}^{(0)}$ with non-zero component along $\mathbf{p}_1$,  the iteration                                                    \begin{equation}                                                                                                                                                    
  \mathbf{u}^{(t+1)}\leftarrow \frac{\mathbf{M}\mathbf{M}^T\mathbf{u}^{(t)}}{\lVert\mathbf{M}\mathbf{M}^T\mathbf{u}^{(t)}\rVert}                                      
  \end{equation}                                                                  
  converges to $\mathbf{p}_1$, the dominant eigenvector of $\mathbf{M}\mathbf{M}^T$, which is exactly the leading left singular vector of $\mathbf{M}$.  Convergence is geometric, with error contracting by a factor of   $(\sigma_2/\sigma_1)^{2}$ per step.

 \paragraph{Computing $\mathbf{M}\mathbf{M}^T\mathbf{u}$ in chunks.} The key property is that the cross-product splits as a sum over modalities: \begin{equation}                                                                                                                                                
  \mathbf{M}_d^{(n)}(\mathbf{V})\,\mathbf{M}_d^{(n)}(\mathbf{V})^T\,\mathbf{u}_d^{(n)}
  =\sum_{m\in\mathcal{M}_d}(\mathbf{X}_{d,m}\mathbf{v}_m^{(n)})\,(\mathbf{X}_{d,m}\mathbf{v}_m^{(n)})^T\,\mathbf{u}_d^{(n)}.                                          
  \end{equation}                                                                   
  Each term $\mathbf{X}_{d,m}\mathbf{v}_m^{(n)}$ is itself computed in row chunks of $\mathbf{X}_{d,m}$, so neither the projection vectors nor any concatenated form is ever held in memory in full. When $|\mathcal{M}_d|$
  is large, the sum can be further restricted at each iteration to a randomly sampled subset of modalities; this introduces noise into the
  estimate of $\mathbf{M}\mathbf{M}^T\mathbf{u}$ but does not bias the limit, and successive power-iteration steps refine the estimate. The same strategy applies to $\mathbf{V}_m$ with the roles of rows and columns swapped.   The pseudo code is summarised in Table~\ref{tab:jointsvd-minibatch}.

  \begin{table}[h!]
  \centering
  \begin{tabular}{l l}
  \hline
  \multicolumn{2}{c}{\textbf{Algorithm: Mini-batch Joint SVD}} \\ \hline
   1: & Initialise $\mathbf{U}_d$, $\mathbf{V}_m$ for $d=1,\dots,D$ and $m=1,\dots,M$;\\
      & set $\mathbf{D}_{d,m}\leftarrow\mathrm{diag}\bigl(\mathbf{U}_d^{T}\mathbf{X}_{d,m}\mathbf{V}_m\bigr)$ for $(d,m)\in\mathcal{M}$ \\                            
      & \textbf{while} not converged \textbf{do} \\
   2: & \hspace{2em} for $(d,m)\in\mathcal{M}$ \\
      & \hspace{2em}\hspace{2em} set $\mathbf{D}_{d,m}\leftarrow \mathrm{diag}\bigl(\mathbf{U}_d^{T}\mathbf{X}_{d,m}\mathbf{V}_m\bigr)$ \\
   3: & \hspace{2em} for each domain $d=1,\dots,D$ \\
      & \hspace{2em}\hspace{2em} for each component $n=1,\dots,r$ \\
  3a: & \hspace{2em}\hspace{2em}\hspace{2em} sample $\mathcal{B}\subseteq\mathcal{M}_d$ of size $\min(|\mathcal{M}_d|,\text{batch\_size})$ \\                         
  3b: & \hspace{2em}\hspace{2em}\hspace{2em} initialise $\mathbf{u}\sim\mathcal{N}(\mathbf{0},\mathbf{I})$, $\mathbf{u}\leftarrow\mathbf{u}/\lVert\mathbf{u}\rVert$ \\
  3c: & \hspace{2em}\hspace{2em}\hspace{2em} \textbf{for} $t=1,\dots,T_{\text{pow}}$ \textbf{do} \quad // power iteration \\
      & \hspace{2em}\hspace{2em}\hspace{2em}\hspace{2em} $\mathbf{w}\leftarrow\mathbf{0}$ \\
      & \hspace{2em}\hspace{2em}\hspace{2em}\hspace{2em} for each $m\in\mathcal{B}$ \\
      & \hspace{2em}\hspace{2em}\hspace{2em}\hspace{2em}\hspace{2em} compute $\mathbf{p}\leftarrow\mathbf{X}_{d,m}\mathbf{v}_m^{(n)}$ in row chunks of size           
  \text{batch\_size} \\
      & \hspace{2em}\hspace{2em}\hspace{2em}\hspace{2em}\hspace{2em} $\mathbf{w}\leftarrow\mathbf{w}+\mathbf{p}\,(\mathbf{p}^{T}\mathbf{u})$ \\
      & \hspace{2em}\hspace{2em}\hspace{2em}\hspace{2em} $\mathbf{u}\leftarrow \mathbf{w}/\lVert\mathbf{w}\rVert$ \quad //
  $\mathbf{w}=\mathbf{M}_d^{(n)}(\mathbf{V})\mathbf{M}_d^{(n)}(\mathbf{V})^{T}\mathbf{u}$ \\
  3d: & \hspace{2em}\hspace{2em}\hspace{2em} $\mathbf{u}_d^{(n)}\leftarrow \mathbf{u}$ \\
  3e: & \hspace{2em}\hspace{2em} $\mathbf{U}_d\leftarrow \mathrm{QR}\bigl([\mathbf{u}_d^{(1)},\dots,\mathbf{u}_d^{(r)}]\bigr)$ \\
   4: & \hspace{2em} for each modality $m=1,\dots,M$ \\
      & \hspace{2em}\hspace{2em} for each component $n=1,\dots,r$ \\
  4a: & \hspace{2em}\hspace{2em}\hspace{2em} sample $\mathcal{B}\subseteq\mathcal{D}_m$ of size $\min(|\mathcal{D}_m|,\text{batch\_size})$ \\                         
  4b: & \hspace{2em}\hspace{2em}\hspace{2em} initialise $\mathbf{v}\sim\mathcal{N}(\mathbf{0},\mathbf{I})$, $\mathbf{v}\leftarrow\mathbf{v}/\lVert\mathbf{v}\rVert$ \\
  4c: & \hspace{2em}\hspace{2em}\hspace{2em} \textbf{for} $t=1,\dots,T_{\text{pow}}$ \textbf{do} \\
      & \hspace{2em}\hspace{2em}\hspace{2em}\hspace{2em} $\mathbf{w}\leftarrow\mathbf{0}$ \\
      & \hspace{2em}\hspace{2em}\hspace{2em}\hspace{2em} for each $d\in\mathcal{B}$ \\
      & \hspace{2em}\hspace{2em}\hspace{2em}\hspace{2em}\hspace{2em} compute $\mathbf{q}\leftarrow\mathbf{X}_{d,m}^{T}\mathbf{u}_d^{(n)}$ in column chunks of size    
  \text{batch\_size} \\
      & \hspace{2em}\hspace{2em}\hspace{2em}\hspace{2em}\hspace{2em} $\mathbf{w}\leftarrow\mathbf{w}+\mathbf{q}\,(\mathbf{q}^{T}\mathbf{v})$ \\
      & \hspace{2em}\hspace{2em}\hspace{2em}\hspace{2em} $\mathbf{v}\leftarrow \mathbf{w}/\lVert\mathbf{w}\rVert$ \\
  4d: & \hspace{2em}\hspace{2em}\hspace{2em} $\mathbf{v}_m^{(n)}\leftarrow \mathbf{v}$ \\
  4e: & \hspace{2em}\hspace{2em} $\mathbf{V}_m\leftarrow \mathrm{QR}\bigl([\mathbf{v}_m^{(1)},\dots,\mathbf{v}_m^{(r)}]\bigr)$ \\
      & \textbf{end while} \\\hline
  \end{tabular}
  \caption{Pseudocode for the mini-batch Joint SVD algorithm. $T_{\text{pow}}$ is the number of power-iteration steps per component (e.g. 2 in our implementation), and \text{batch\_size} controls both the number of sampled matrices and the chunk size for matrix-vector products. $\mathcal{D}_m$ denotes the set of domains for which modality $m$ is observed.}                                                                                                                                    
  \label{tab:jointsvd-minibatch}                                                                                                                                      
  \end{table}

\begin{figure}[h]
    \centering
    \includegraphics[width=0.8\linewidth]{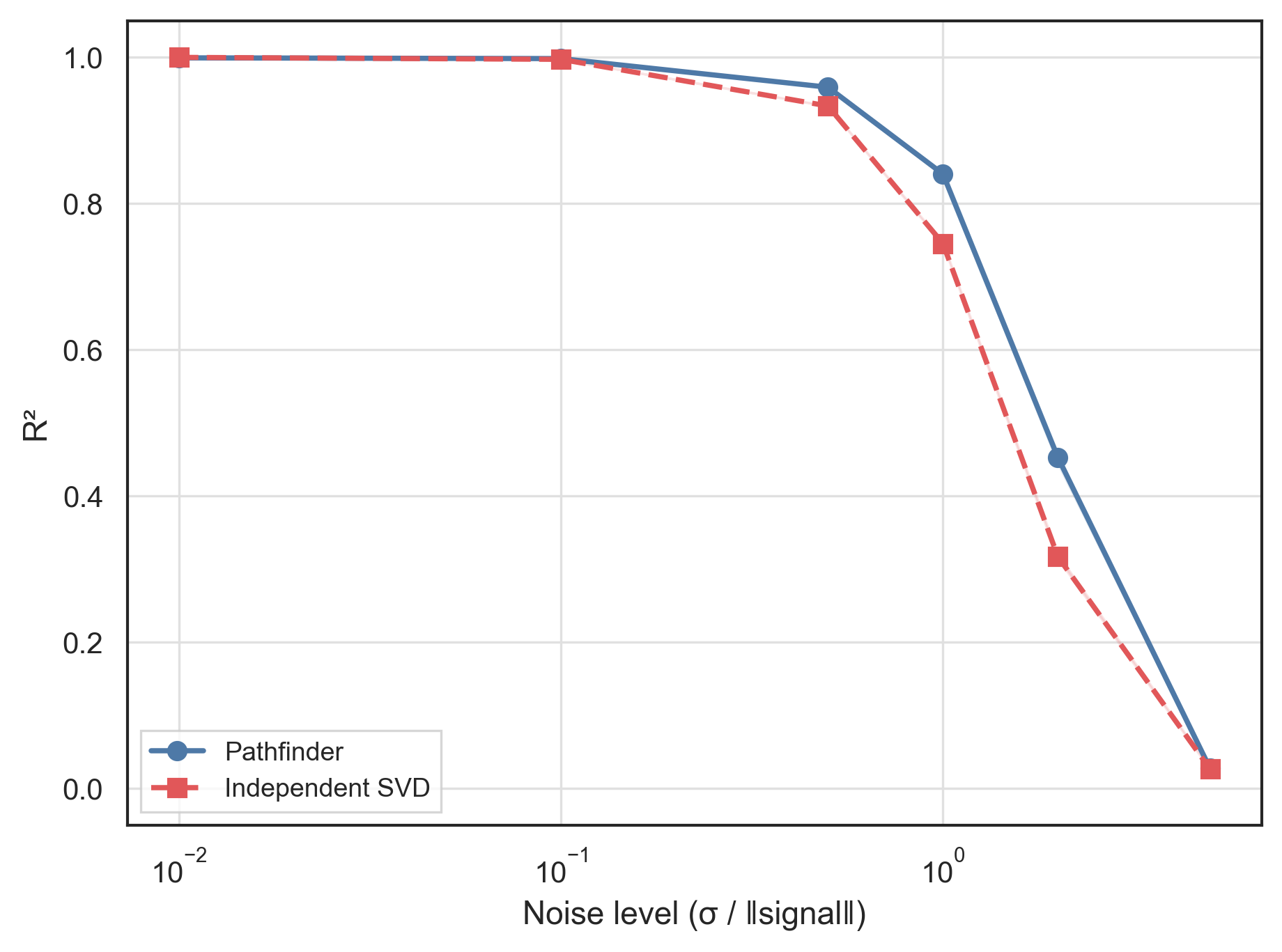}
    \caption{\textbf{Robustness to noise of PathFinder and independent SVD}. PathFinder gave higher accuracy when the noise level is high, suggesting a denoising effect by sharing factors across domains and modalities. }
    \label{fig:svd}
\end{figure}

\begin{figure}[h]
    \centering
    \includegraphics[width=\linewidth]{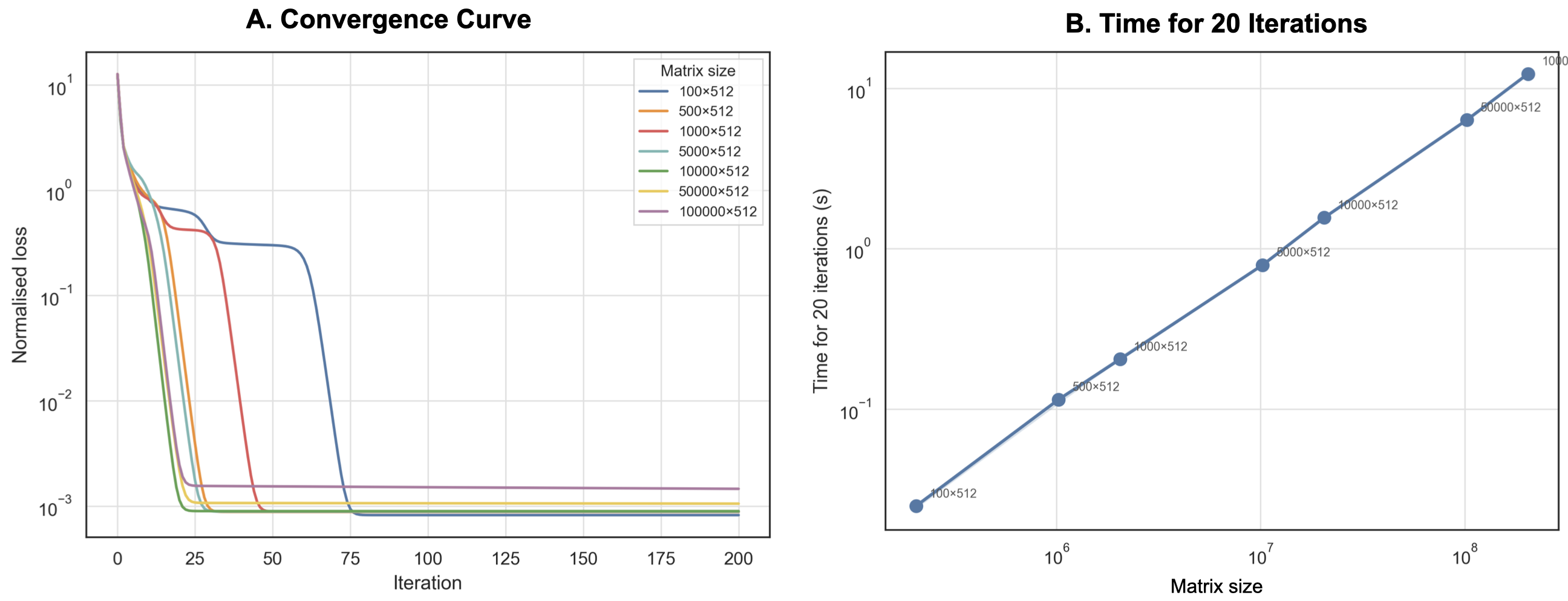}
    \caption{\textbf{(A)} Convergence curve of various matrix sizes. Larger matrix size gives faster convergence. \textbf{(B)} Time required for 20 iterations as a function of matrix size.}
    \label{fig:convergence}
\end{figure}

Figure \ref{fig:svd} examines the robustness to noisy input in the joint SVD approach. Figure \ref{fig:convergence} reports on the convergence of the mini-batch algorithm as a function of input data size. 

\end{document}